\documentclass[10pt,twocolumn,letterpaper]{article}
\def\abstract 
    {
    \centerline{\large\bf Abstract}%
    \vspace*{12pt}%
    \it%
    }
 
\usepackage{enumitem}
\setlist[itemize]{noitemsep,leftmargin=*,topsep=0em}
\setlist[enumerate]{noitemsep,leftmargin=*,topsep=0em}
\usepackage{graphicx}
\usepackage{amsmath}
\usepackage{amssymb}
\usepackage{booktabs}

\usepackage{times}
\usepackage{epsfig}
\usepackage{graphicx}
\usepackage{xcolor}
\usepackage{caption}
\usepackage{subcaption}
\usepackage{array}
\usepackage{multirow}
\usepackage[numbers,sort&compress]{natbib}
\makeatletter
\newcommand\extralabel[2]{{\edef\@currentlabel{\@currentlabel#2}\label{#1}}}
\makeatother

\usepackage[breaklinks,colorlinks]{hyperref}

\usepackage[capitalize]{cleveref}

\title{Unveiling Ancient Maya Settlements Using Aerial LiDAR Image Segmentation}

\author{Jincheng Zhang$^1$, William Ringle$^2$, and Andrew Willis$^1$\\
$^1$University of North Carolina at Charlotte, Charlotte, NC 28223 USA\\
$^2$Davidson College, Davidson, North Carolina, USA\\
{$^1$\tt\small \{jzhang72, arwills\}@charlotte.edu} \\
{$^2$\tt\small biringle@davidson.edu}
}

\date{} % Comment this line to show today's date

\begin{document}

\maketitle

\begin{abstract}
%The labor-intensive task of manually identifying archaeological structures in LiDAR imagery poses a significant challenge for archaeologists, demanding meticulous effort and time-consuming analyses. Fortunately, advancements in deep learning (DL) offer new perspectives on archaeological research by providing efficient solutions for automating the identification of historical features. This paper presents a pioneering application of the YOLOv8 convolutional neural network for the segmentation of archaeological structures in aerial LiDAR images. YOLOv8 is adapted and fine-tuned to discern and delineate archaeological structures with high precision and efficiency. Our methodology involves pre-processing raw LiDAR data, extracting relevant features, and training the YOLOv8 network on a curated dataset of archaeological structures. The segmentation results for two classes, annular structures, and platforms, showcase the efficacy of YOLOv8 in accurately identifying and delineating archaeological structures from aerial LiDAR imagery, with an IoU score of 70\% on a LiDAR survey of 478$km^2$. The archaeological analysis in this article explores the potential of YOLOv8 as a powerful tool for automated segmentation in aerial LiDAR images and paving the way for more efficient exploration and preservation of historical landscapes. The proposed approach not only enhances the speed of analysis but also exhibits promising accuracy, making it a valuable tool for archaeologists and researchers in uncovering hidden remnants of the past.}
Manual identification of archaeological features in LiDAR imagery is labor-intensive, costly, and requires archaeological expertise. This paper shows how recent advancements in deep learning (DL) present efficient solutions for accurately segmenting archaeological structures in aerial LiDAR images using the YOLOv8 neural network. The proposed approach uses novel pre-processing of the raw LiDAR data and dataset augmentation methods to produce trained YOLOv8 networks to improve accuracy, precision, and recall for the segmentation of two important Maya structure types: annular structures and platforms. The results show an IoU performance of 0.842 for platforms and 0.809 for annular structures which outperform existing approaches. Further, analysis via domain experts considers the topological consistency of segmented regions and performance vs. area providing important insights. The approach automates time-consuming LiDAR image labeling which significantly accelerates accurate analysis of historical landscapes.

\end{abstract}

\section{Introduction\label{sec:intro}}

\begin{figure*}[t]
  \centering
  \begin{subfigure}[t]{0.24\textwidth}
  \includegraphics[width=\linewidth, height=1.062\linewidth]{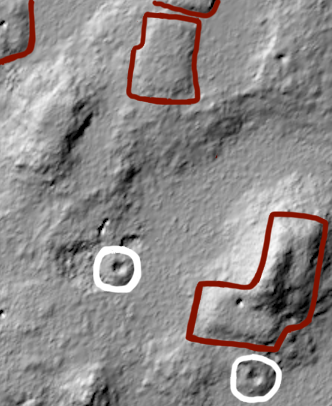} 
   \caption{Platforms and annular structures. }
   \label{fig:platform_and_annular}
  \end{subfigure}
  \hfill
  \begin{subfigure}[t]{0.37\textwidth}
  \includegraphics[width=\linewidth, height=0.695\linewidth]{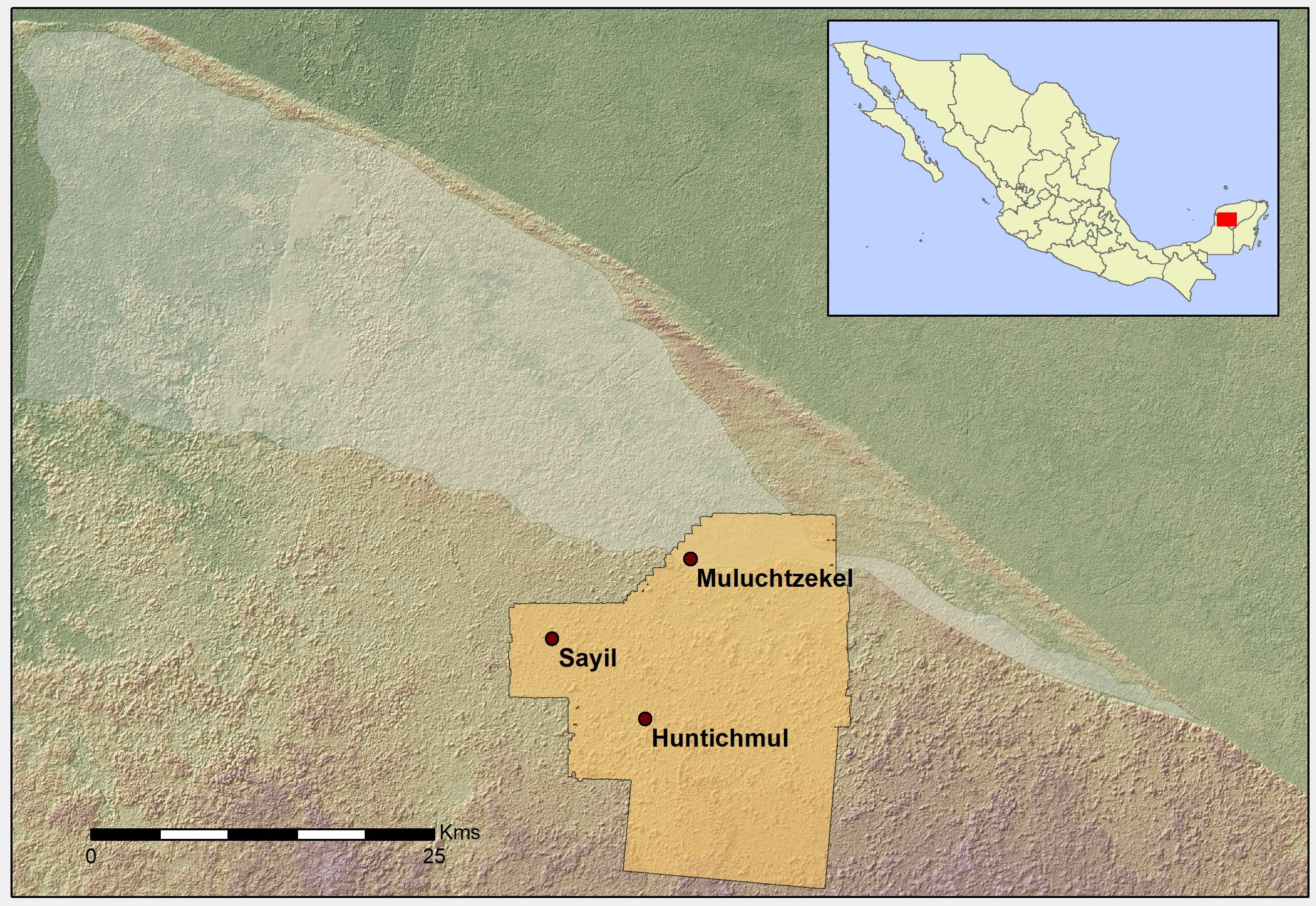} 
   \caption{Location map of the Puuc region of Yucatan, Mexico. }
   \label{fig:research_map}
  \end{subfigure}
  \hfill
  \begin{subfigure}[t]{0.37\textwidth}
    \includegraphics[width=\linewidth, height=0.69\linewidth]{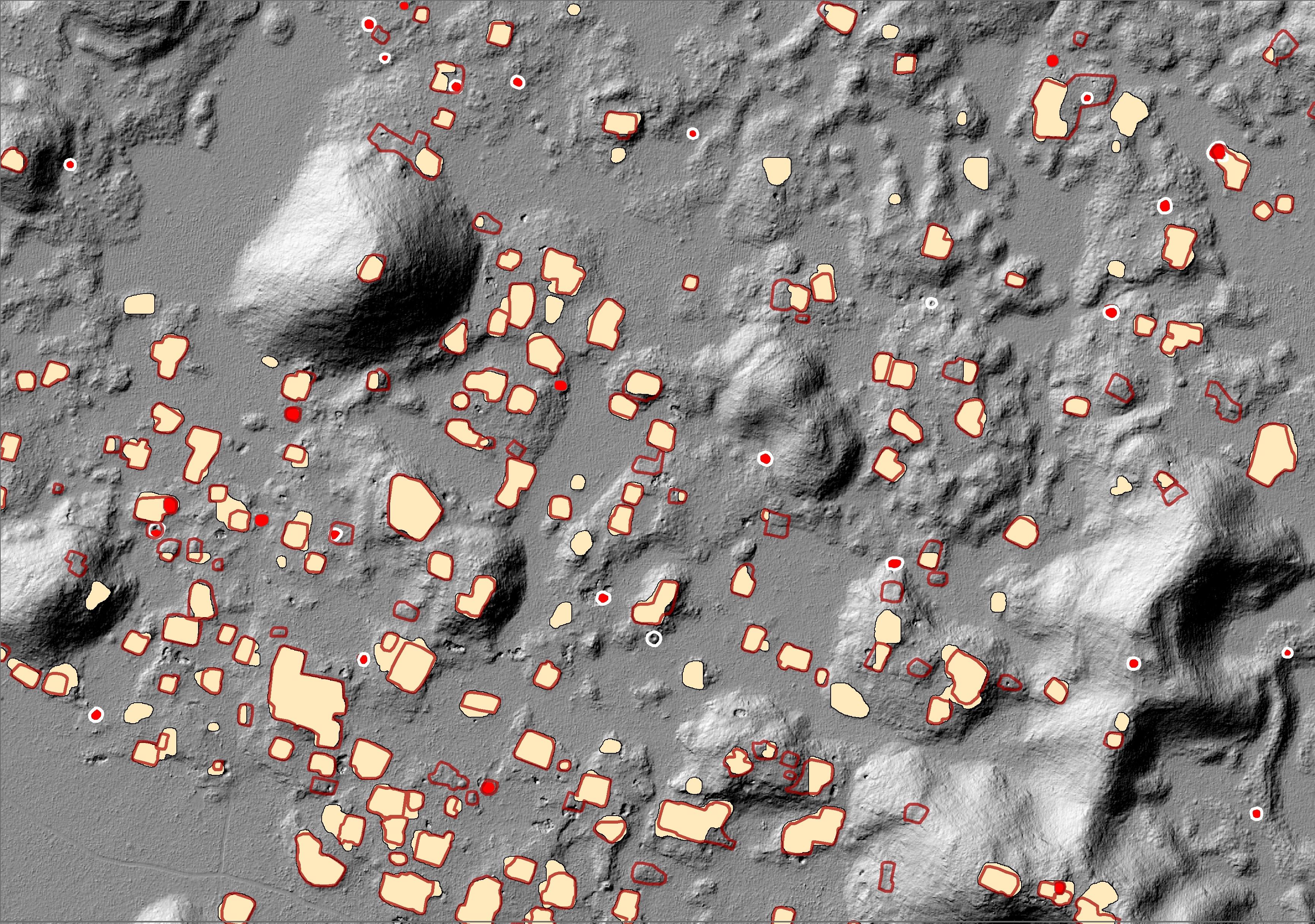}
  \caption{Inference result on the Huntichmul site. }
  \label{fig:HNT_seg_result}
  \end{subfigure}
  \label{fig:sfsd}
  \caption{(a) Examples of platforms and annular structures visualized in a hillshading image where platforms are outlined in brown and annular structures are in white. (b) Maya structures found in three sites, Muluchtzekel (MLS), Sayil (SAY), and Huntichmul (HNT), in the Puuc region of Yucatan, Mexico were used as data for analysis. (c) The inference result of Huntichmul demonstrates the proposed system's capability of extracting platforms and annular structures. The ground truth and predicted masks were superimposed onto a hillshading image for better visualization. The brown and white polygons respectively outline the ground truth boundaries of platforms and annular structures, while the solid beige and red regions show the predictions of YOLOv8 respectively.}
\end{figure*}

Settlement archaeology attempts to determine the population of a past society and its disposition with respect to its surroundings. Sufficiently large samples of past landscapes have often been difficult to obtain because of dense vegetation and/or rugged terrain. However, starting in 2008, the introduction of LiDAR remote sensing technology to Maya archaeology allowed aerial coverage of hundreds of square kilometers within a few days \cite{chase2011airborne,ringle2021lidar,canuto2018ancient,yaeger2016locating,vsprajc2022archaeological,stanton2020structure}. As a result, archaeologists now often face a surfeit of data, demanding the investment of many hours of skilled analysis. For example, over 60,000 house platforms were detected in LiDAR coverage of a large campaign across the lowlands of Guatemala \cite{canuto2018ancient}. As LiDAR coverage of the Maya lowlands continues to expand, the burden of image analysis will increase dramatically, since the ancient population is estimated to have ranged into the millions.

%\br{In this report we discuss our efforts to apply deep learning technologies to the task of feature recognition in a large LiDAR dataset from the Puuc region of Yucatan. The Puuc was one of the most densely settled areas of the ancient Maya between AD 700-950 and thereafter was largely abandoned. Our focus is the identification of two classes: (1) stone platforms which supported houses and other buildings and (2) annular structures, circular donut-shaped mounds which are probably the remains of ancient pit kilns used for lime production. The latter are usually about 10-12 m in diameter, but stone platforms vary from a few meters on a side to well over a hundred meters, though most range from 200-1200 m2. }

The development of deep learning technologies has opened up vast possibilities for enhancing Maya settlement recognition within large LiDAR datasets, particularly through rapid object detection and segmentation techniques. Deep learning models can efficiently identify, classify, and segment objects of interest with high accuracy and speed. By leveraging these technologies, researchers can streamline the process of analyzing vast LiDAR datasets, enabling comprehensive surveys of Maya archaeological sites at a scale previously unattainable \cite{somrak2020learning, bundzel2020semantic, ayala2022deep, jannat2023extracting, hellweg2022ensemble}. Moreover, deep learning facilitates the discovery of previously unnoticed objects and features, enriching the understanding of Maya civilization and cultural heritage. 
%As these technologies continue to advance, their potential to revolutionize Maya object recognition in large LiDAR datasets is increasingly evident, promising new insights and discoveries that contribute to the broader field of archaeology.

%\br{We previously used neural networks to identify archaeological features using image classification, as had other projects \cite{seligson2017using}. Although promising, this method only provides rectangles around identified features. In contrast, we here report experiments with YOLOv8 image segmentation, allowing not only the detection of features but also the determination of their spatial extension. Figure 2 presents a hillshade image of Huntichmul, a site previously covered by a ground survey that we used as a test case. Figure 2b is a visualization of the test results, in which the outlines of platforms verified by ground survey are given in blue and the YOLOv8 predictions are filled in pink. Accurate knowledge of the position and shape of features in turn can potentially be used to derive their height and volume, as well as their disposition with respect to neighboring features. }

This study explores the application of the state-of-the-art DL network YOLOv8 \cite{JocherUltralyticsYOLO2023} for segmenting Maya structures found in the Puuc region of Yucatan, Mexico. The research focuses on two classes: (1) stone platforms supporting houses and other buildings, and (2) annular structures, circular mounds likely remnants of ancient pit kilns for lime production. Fig. \ref{fig:platform_and_annular} illustrated some examples of platforms and annular structures. Annular structures are usually about 10-12 $m$ in diameter while stone platforms vary in size from a few meters to over a hundred meters, with most ranging from 200-1200 $m^2$. Training and evaluation of the deep learning system were conducted using objects of interest from three Puuc sites, Muluchtzekel (MLS), Sayil (SAY), and Huntichmul (HNT), as depicted in Fig. \ref{fig:research_map}. The LiDAR data of these sites were collected by the National Center for Airborne LiDAR and Mapping (NCALM) using a Teledyne Optech TitanMW(14SEN/CON340) sensor mounted in an airplane flying at an altitude of 600-650 $m$. From this, a 0.5 $m$ resolution Digital Terrain Model (DTM) raster was produced where each pixel measures 0.5 meters in both width and height. To train YOLOv8 for Maya structure segmentation, LiDAR data were processed to generate image datasets of $256\times256$ pixels, using objects manually labeled and verified by ground surveying during the years 1984 - 2023.

The contributions of this article are:

% \begin{itemize}
%     \item Identification of effective representations of the Aerial LiDAR data to improve the segmentation results. These results include archaeological evaluation metrics such as region topology making the result analysis more rigorous than typically published.
% %    \item Our transformations of the measured data yields improved accuracy archaeological structures in aerial LiDAR images, showcasing its high accuracy (0.842 IoU for platforms and 0.809 for annular structures).
%     \item Development of novel pre-processing of the raw LiDAR data and dataset augmentation methods that preserve accuracy with a small training dataset.
%     \item Proposal of a multi-scale inference approach and novel post-processing methods to improve the performance.
%     \item Demonstration of performance across multiple archaeological sites.
%     \item Discussion of end-user benefits of the system by a domain expert, detailing how the technology impacts large-scale survey annotation.
%     \item Creation of ablation studies examining the impact of (1) several pre-processing filters and (2) image scaling which can dramatically impact performance.
% \end{itemize}
%\br {It's not wrong as is, but I think this would read a little better if nouns replaced the gerunds, i.e. Identification for identifying, Development for developing, etc.}

\begin{itemize}
     \item Identification of effective representations of aerial LiDAR data that best enhance segmentation results, incorporating archaeological evaluation metrics such as region topology for a more rigorous analysis than typically published.
     \item Development of novel pre-processing techniques for raw LiDAR data and dataset augmentation methods that maintain accuracy with a small training dataset.
     \item Introduction of a multi-scale inference approach and novel post-processing methods to improve the segmentation performance.
     \item Demonstration of performance across multiple archaeological sites, providing a comprehensive evaluation of the proposed methodology's effectiveness.
     \item Discussion of end-user benefits of the system by a domain expert, detailing how the technology positively impacts large-scale survey annotation.
     \item Ablation studies examining the impact of different data representations and image scaling on segmentation performance, offering insights into key influencing factors.
\end{itemize}

%\br{This article navigates through the current landscape of deep learning applications in archaeology, contextualizing the utilization of YOLOv8 within the broader framework of Maya archaeology. We discuss the intricacies of adapting and fine-tuning YOLOv8 for the specific requirements of ancient Maya structure segmentation. Through case studies and comparative analyses, this paper explores the advantages and challenges posed by YOLOv8 in enhancing the accuracy and speed of segmentation processes.}

%The demonstrated accuracy underscores YOLOv8's potential in handling archaeological landscapes efficiently. The proposed dataset generation pipeline facilitates efficient model training by creating training data from a few high-resolution aerial LiDAR scans. The investigation into YOLOv8's generalizability across diverse archaeological sites was conducted by applying the trained model to a new site outside the sites containing the training and testing samples, showcasing its broader applicability beyond specific training datasets. Through case studies and comparative analyses, this paper highlights how automated identification can streamline large-scale surveys, enabling archaeologists to focus on higher-level analysis. Finally, the detailed ablation study identifies optimal image visualizations and data augmentation parameters for semantic segmentation tasks. These contributions offer valuable insights into leveraging YOLOv8 within the context of Maya archaeology.

The result of these contributions is the creation of a new system that automates the identification and segmentation of archaeological annular structures and platforms, demonstrating the immense potential of deep learning in streamlining labor-intensive tasks traditionally reliant on manual expertise.

\begin{table*}[ht]
\centering
\begin{tabular}{cccccccc}
\hline
    Site  & Overall Area ($km^2$) & Area Surveyed ($km^2$) & Hills (\%) & Flats (\%) & Low Rises (\%)  & Platforms & Annulars \\ 
\hline
\hline
    MLS   & 6.00    &   4.03 & 0.172 & 0.081 & 0.747  & 564 & 61   \\ 
    % Muluchtzekel Block 1 & 600.00                & 56.0                         & 110                &                   \\ \hline
    % Muluchtzekel Block 2 & 600.00                & 347.0                        & 503                &                   \\ \hline
    %Kom   & 1.25   & 0.907     & --     & 32   \\ 
    %Uchbenmul   & 1.09   & 0.311    & --       & 10      \\ 
    SAY & 12.00    &    3.14  & 0.127 &   0.021 & 0.852  &  646   &  112   \\
    HNT &4.50    &    1.04 & 0.243  & 0.061 & 0.696 &  513  & 70   \\
\hline
\end{tabular}
\caption{\label{tab:tab-gnd-surv}Ground survey statistics for three archaeological sites used in this paper. Areal statistics are followed by the distribution of terrain types, revealing that the landscapes of MLS and SAY have fewer hills when compared to HNT. ``Hills'' have a local height of 30 $m$ or more, ``Flats'' have virtually no relief, and ``Low Rises'' comprise the remaining terrain. The last two columns show the number of manually identified platforms and annular structures.}
\end{table*}

\section{Related Work}

This section summarizes common data representations in imaging Maya structures, followed by recent developments in the segmentation of ancient Maya structures using deep learning networks.

\subsection{ALS Raster Representations}

%Airborne Laser Scanning (ALS) is a powerful technique used in archaeology to create detailed maps and 3D models of terrain features using laser pulses (LiDAR) from an airborne platform. Raw ALS data is often converted to Digital Elevation Models (DEMs) which is a digital representation of surface elevation data to help archaeologist inspect the landscape. ALS raster visualizations are grayscale or color raster images converted from the numerical DEM data. Archaeologists also leverage these visualizations to identify subtle terrain variations and archaeological features such as riverbanks, cultural terraces, stone walls, or areas of erosion \cite{kokalj2017airborne}.
Airborne Laser Scanning (ALS) is a powerful technique used in archaeology to create detailed 3D terrain models using laser pulses (LiDAR) gathered by an airborne sensor. For archaeological purposes, raw ALS point data are often converted to "bare earth" topographic models referred to as Digital Terrain Models (DTM). ALS raster data representations are the grayscale or color raster image representations derived from the DTM.%, for example, elevation, and slope. Archaeologists can leverage these visualizations to identify subtle terrain variations and archaeological features such as structures, riverbanks, cultural terraces, stone walls, or areas of erosion \cite{kokalj2017airborne}.

%\br {The ALS datasets used herein derive from unlabeled LiDAR data collected by the National Center for Airborne LiDAR and Mapping (NCALM) over the Puuc region of Mexico in Fig. \ref{fig:reasearch_map}. Data collection was sponsored by NSF Grant 166503 award to the Bolonchen Regional Archaeological Project and is used with their permission. Lidar was obtained using a Teledyne Optech TitanMW (14SEN/CON340) sensor mounted in a small airplane flying at an altitude of 600-650 m. About 62.6\% of the pulses produced ground returns, resulting in a density of 10.6/m2 over the entire 478 $km^2$ of coverage. From this, a 0.5 m DTM raster was produced, forming the input for generating the SPS images described in Sec. \ref{sec:data_gen}. Preliminary inspection indicates the region contains approximately 14,000 platforms and 2,100 annular structures.}

ALS raster data representations \cite{kokalj2017airborne}, including hillshading (HS), slope, sky-view factor (SVF), openness, etc, offer contrasting insights into terrain characteristics. HS provides a visually intuitive representation based on the assumption of a Lambertian surface (equally bright from all directions) \cite{oren1994generalization} illuminated by a distant light source. The slope is the first derivative (gradient) of the surface elevation. SVF depicts the visible sky portion from a point, offering an alternative to hillshading to overcome the directional problem discussed below. Openness estimates horizon elevation angles within a defined radius, with positive openness (PO) being the mean zenith angle value. %Fig.\ref{fig:als_vis} shows an example image for each ALS visualization. 

Combining representations can provide additional insights. Visualization for Archaeological Topography (VAT) is a widely used method, blending HS, slope, PO, and SVF representations. While VAT effectively shows the topographic features of archaeological sites and landscapes, its reliance upon hillshading makes it orientation-dependent. This limits its compatibility with data augmentation techniques like rotation and flipping \cite{kokalj2023machine}, which are crucial for data processing in deep learning. An alternative combination was proposed in \cite{somrak2020learning}, creating a three-channel image representation using SVF, PO, and slope for DL classification tasks. This paper applies this representation to semantic segmentation and evaluates its effectiveness.
%A combination of visualizations can provide insights from different perspectives. Visualization for Archaeological Topography (VAT) is a widely used method, blending HS, slope, PO, and SVF visualizations. While VAT effectively showcases the topographic features of archaeological sites and landscapes, it relies on hillshading, making it orientation-dependent. This limits its compatibility with data augmentation techniques like rotation and flipping, crucial for deep learning. An alternative combination was proposed in \cite{kokalj2023machine}, creating three-channel (RGB) image representations using SVF, PO, and slope for deep-learning classification tasks. This paper extended this visualization to the context of semantic segmentation and presented its effectiveness.

\subsection{Segmentation on Archaeological Structures}

Deep learning has emerged as a powerful tool for analyzing LiDAR and aerial imagery datasets in archaeological research, particularly in tasks like classification and semantic segmentation \cite{somrak2020learning, bundzel2020semantic, banasiak2022semantic, kocev2022discover, jannat2023extracting, hellweg2022ensemble}. Semantic segmentation involves accurately delineating architectural features within these datasets. Recent studies, such as those utilizing the U-Net architecture \cite{ronneberger2015u}, have shown promising results, especially in identifying ancient Maya settlement features \cite{ayala2022deep, bundzel2020semantic, jannat2023extracting}. Other architectures like DeepLabv3 \cite{chen2017rethinking} and HRNet \cite{wang2020deep} have also been explored in \cite{kocev2022discover}. %These studies demonstrate the potential of semantic segmentation in identifying Maya structures and enhancing archaeological analyses. 

This article demonstrates methods that enable accurate segmentation of Maya platforms and annular structures using YOLOv8. To our knowledge, this neural architecture has not been applied to this application domain. The proposed methodology and results indicate the pre-processing steps needed to make YOLOv8 significantly outperform competing neural models.
%The results highlight the high precision and efficiency of YOLOv8 in the task of Maya structure segmentation.
%https://www.researchgate.net/publication/354696755_Ensemble_Learning_for_Semantic_Segmentation_of_Ancient_Maya_Architectures
%https://www.mdpi.com/2072-4292/14/4/995
%https://arxiv.org/pdf/2208.03163.pdf
%https://arxiv.org/pdf/2208.03163.pdf#page=15

\section{Methodology}

This section describes the DL model, how the data were conditioned for training the model, and how the trained model was applied to process large raw LiDAR data. The discussion is organized into the following sections:

\begin{itemize}
    \item Data pre-processing steps that modify raw LiDAR data into measurements that yield higher performance.
    \item Data augmentation steps that expand the training database and yield higher performance.
    \item Inference processing methods that allow very large LiDAR images, for example, 3600$\times$5000, to be processed using a network having a 256x256 input layer and coping with tiling artifacts in the labeling.
    \item A brief overview of the DL model, YOLOv8.
    \item A summary of the evaluation approach including conventional pixel-based metrics and novel object-based topological metrics.
\end{itemize}

%Customized data augmentation approaches were applied to generate the image dataset for training and testing the YOLOv8 segmentation network \cite{JocherUltralyticsYOLO2023}. Metrics, including Intersection of Union (IoU), precision, and recall rate, were used to evaluate the segmentation performance. To infer new image data using the trained weights, the proposed method performed segmentation on small image tiles and merged multiple segmentation results for each pixel into a final segmentation result followed by post-processing.

\subsection{Dataset Pre-processing And Ground Truth \label{sec:data_gen}}

%\jz{insert this: Unlike previous research where these two structures were combined into a single dataset \cite{jannat2023extracting, kocev2022discover, kokalj2023machine}, despite platforms typically outnumbering annular structures significantly, the proposed approach addressed the data imbalance problem by generating two separate datasets to study them individually. }
Table \ref{tab:tab-gnd-surv} presents statistics of the measured data and ground survey data for the three Puuc sites indicated in Fig. \ref{fig:research_map}.  Significantly more platforms than annular structures are present in all three sites, with the former outnumbering the latter by approximately 6 to 9 times. This discrepancy in numbers can lead to data imbalance issues, where models trained on such datasets may exhibit biases favoring the majority class and consequently perform inadequately on minority classes. To address this concern, two distinct image datasets were generated: one specifically for platforms and another for annular structures.

The image for training and testing the network consists of three image channels: (1) \textbf{S}VF, (2) \textbf{P}O, and (3) \textbf{S}lope which we refer to as an SPS image (from the first letters of each channel) similar to work in \cite{somrak2020learning}. The SPS image has the same resolution as the LiDAR DTM where each pixel spans 0.5 meters in width and 0.5 meters in height. This choice was motivated by the ablation study results which showed that SPS images yield better performance than other representations. Fig. \ref{fig:als_vis} illustrated an SPS image tile and other ALS representations including SVF, PO, and slope.% shows a tile imaged by each of these methods (SPS, SVF, PO, slope, HS, elevation).%shows an example of SPS images and associated SVF, PO, and slope images.

%\jz{needs Dr. Ringle to check}The ground truth labels of objects of interest for developing deep learning (DL) systems were generated using the georeferenced survey map for each archaeological site. Structure locations were used to visually trace the outlines of objects, for example, platforms and annular structures. The ground truth labels were then specified manually as polygons that enclose each object. A small subset of structures (\jz{percentage?}) were invisible in the LiDAR imagery and were excluded in the ground truth labeling on the logic that structures identified by DL systems but invisible in the LiDAR imagery would be impossible to verify by archaeologists without further fieldwork.

The labeled SPS images, with sizes up to 6000$\times$8000, were divided into small tiles of 256$\times$256 to generate the dataset for DL systems. Specifically, an image tile was created at the location of each object of interest with the tile center overlapping the object center. 

Compared to image datasets of contemporary objects, the availability of ground-verified ancient Maya structures is extremely limited, and collecting a large labeled dataset of them is challenging and expensive, hence the small sample sizes in Table \ref{tab:tab-gnd-surv}. The ground truth labels of objects of interest were generated by superimposing georeferenced survey maps created between 1984 and 2023 on the LiDAR imagery (SAY data from \cite{sayilmap}, MLS and HNT by a domain expert). These were used to trace the outlines of house platforms and annular structures. Since mapping did not completely cover the regions of interest, objects of interest outside the surveyed area were traced manually using LiDAR imagery to provide sufficient data for training a deep learning system. Small structures invisible in the LiDAR imagery were excluded from ground truth labeling on the logic that similar structures identified by DL systems but invisible in the LiDAR imagery would be impossible to verify without further fieldwork. The ground truth labels were then exported as polygons enclosing each object. 

\begin{figure}
    \centering
    \includegraphics[width=0.8\linewidth]{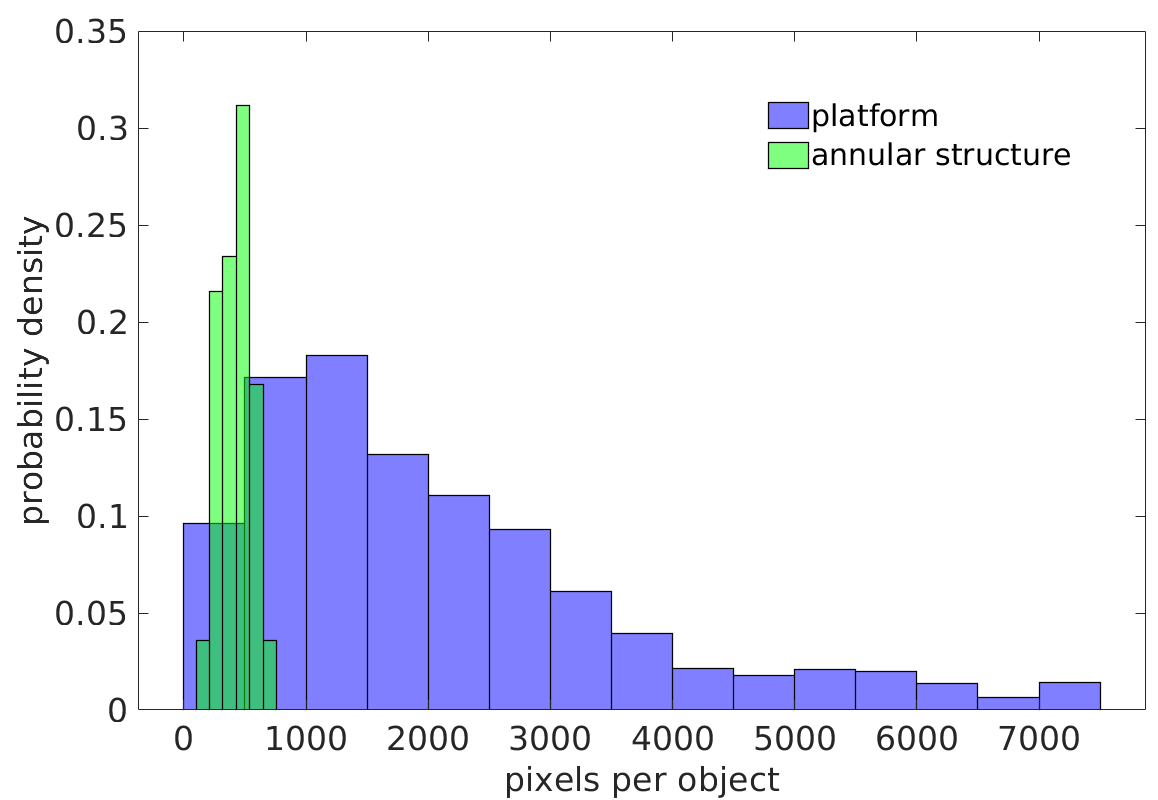}
    \caption{Pixel count per object for platforms and annular structures from the LiDAR images used in this paper.}
    \label{fig:hist_area}
\end{figure}

\subsection{Data Augmentation}
\label{sec:data_aug}

The datasets present significant challenges for ML systems due to two attributes: (1) the small size of the label sets, and (2) the small size of labeled regions. 

Due to the small size of the label sets, data augmentation was required to boost performance. The proposed augmentation approach adapted the methods in \cite{jannat2023extracting}, including (1) random background sampling that collects random samples of background unrelated to the location of labeled objects, and (2) random rotations and translations of each training object. 
These methods virtually expand the dataset and improve data variation. However, translation during augmentation can move segmented structure regions to the image periphery and clip out significant portions of the structure data. When there is a very small extent of the structure in the augmented image, i.e., less than 10 pixels, the inclusion of the region label data adversely impacts segmentation since these data do not have sufficient pixels that can represent the features of objects. To address this, structures that are clipped and possess bounding boxes smaller than 10 pixels in either width or height have their labels removed.

%The resizing augmentation proposed here contrasts with the resizing augmentation typically employed in the training of deep learning systems.
Augmentation of the datasets also included anisotropic scaling that scaled only $(X, Y)$ dimensions of images by 2. Scaling the image can enlarge small objects and improve DL model performance. Fig.~\ref{fig:hist_area} shows the pixel count per object respectively for platforms and annular structures in the image data. Most objects are smaller than 3000 pixels, occupying less than 4.5\% of a 256$\times$256 image tile. This anisotropic approach is important to preserve the height ($Z$ dimension) of the archaeological objects relative to their surroundings and the heights of objects, e.g., buildings, on platforms that do not typically scale with the platform $(X, Y)$ footprint. Unlike conventional scaling approaches, we scale the SPS image at its captured resolution (on the order of 6000$\times$8000) before dividing it into tiles, which has the impact of varying the perceptual field of each object. 

%boundaries with 10 pixels or fewer have their labels removed.

%The second modification omits the object instances that only have a small portion inside the tile perceptual field after the random rotation and translation. These instances do not have sufficient pixels within the image tile to provide the features of objects and, thus are not considered as training samples.  This approach aimed to mitigate the risk of overfitting resulting from limited data variance.

% Two additional modifications were made to provide more distinct representations of data and enhance the DL model performance: (3) resizing the SPS image by two and dividing both the original and resized images into tiles to generate training and testing data, and (4) excluding ground truth labels for objects that are on the boundary of a tile and have bounding boxes smaller than 10 pixels in either direction. %Resizing images increases the number of pixels available for analysis, helping the deep learning model capture finer details and extract more informative features from small objects. 

Due to the diverse shapes of platforms compared to the mostly circular annular structures, which remain unchanged with rotation, the proposed pipeline generated 15 augmented variations at each scale for each platform object and 10 for each annular structure. The final realization of the augmented platform dataset comprises a total of 33,971 images, each sized at 256$\times$256 pixels, while the annular structure dataset contains 3,216 images of the same dimensions. 

\subsection{Multi-Scale Inference and Post Processing\label{sec:infer}} 
%For each unlabeled SPS image of a new archaeological location, 

%\jz{I reorganized this section a little bit and renamed the title to call back a new contribution I added in the introduction section.}
%\br{I think we should refer to resolution as .5 m, rather than .25 m2. It is possible to have other xy dimensions resulting in .25 m2 and .5 m is how resolution is usually specified in the literature I read. Of course you may sometimes need to refer to the area, but .5m resolution will be the more familiar usage. And does a magnification of 2 halve the pixel area?}

The proposed multi-scale approach performs inference on LiDAR image at its native resolution, i.e., each pixel is 0.25 $m^2$, and a magnification of 2, i.e., each pixel is 0.125 $m^2$. The magnified results are scaled down to the native resolution preserving positively labeled $(X, Y)$ locations and merged with the lower resolution region using a logical OR of the results. 

The final inference results are generated by post-processing the output from YOLOv8 to address two issues: (1) tiling artifacts created by partitioning very large raw LiDAR images in tiles for processing, and (2) removal of small segmented regions that are not likely to be of archaeological interest. 

%datasets require the raw image to be divided into tiles having size compatible with the input layer of the DL model. Standard tiling approaches produce tiling artifacts which incorrectly generate linear boundaries to object regions.

%The proposed inference method differs from the standard approach for DL model inference. 
Tiling artifacts occur when structure boundaries are clipped before being input to the DL network. Segmentation for clipped objects often has incorrect linear boundaries that follow the tile boundary. As other researchers have done, a sliding window inference was applied for the image data which classifies the pixels multiple times for different window offsets. Each classification generates a segmentation mask and the logical OR of the generated masks across all windows generates the final segmentation result.

%resized image by scaling the original image by 2, then decomposes both the original image and resized image into 256$\times$256 tiles where these tiles may overlap. The inference results of each pixel in each tile are merged via a logical OR of the masks to generate a final segmentation. Results of this article reflect a sliding window of 80 pixels in both $(X,Y)$ directions.

%This method, referred to as multi-scaled sliding-window inference, allows the inferred output to (1) improve the segmentation of small objects, (2) take into consideration image values from a perceptual field larger than the tile, and (3) reduce adverse impacts associated with labels that lie on the boundary of the image. The sliding-window inference in this paper considers sliding window skips of 80 pixels horizontally and vertically. For the chosen tile size, this results in inferring most pixels (except for the pixels on image corners) 16 times. 

A review of the ground truth label set reveals that platforms and annular structures have well-defined minimum area limits, 41 $m^2$ and 33 $m^2$ respectively, and only a small number of objects have areas close to these minimum thresholds. The proposed post-processing implements this prior knowledge to remove spurious small regions. Specifically, segmented output with bounding boxes smaller than 15 pixels (7.5 $m$) in either $X$ or $Y$ dimension were filtered out. This was achieved by a series of standard morphological operations.

%\br {I don't know if we can really defend this - it seems like something a reviewer would jump on. Maybe just say that after experimenting with a variety of filters, we found 15x15 optimally eliminated spurious segments while recognizing actual features}

%Due to the limited amount of training data, the trained DL model creates some very small segmentation results when inferring a new image. Analysis indicates that most of those small results are small natural features in the output. A post-processing approach was designed to remove them and improve the inference performance. Specifically, segmented outputs that are smaller than 15$\times$15 pixels were filtered out. This was achieved by a morphological opening operation followed by another operation that removes small connected components from the binary mask based on the specified size criterion. 
%Due to the limited amount of training data and the small size of interested objects, the trained DL model tends to create some very small segmentation results when inferring a new image. Analysis from the archaeologist in this project indicates that most of those small results are the \textit{noise} in the output. 
%This post-processing approach was designed to remove the noise and improve the inference performance of the trained DL model.

\subsection{YOLOv8 Deep Learning Architecture}

%YOLOv8 was adapted to perform the segmentation task in this paper. 
YOLOv8 is the latest state-of-the-art YOLO (You Only Look Once) model that can be used for object detection, image classification, and instance segmentation tasks. The YOLOv8 segmentation model, as an extension of its detection model, makes use of a few key components to enhance the performance including (1) a CSPDarknet53 \cite{wang2020cspnet} feature extractor as the model backbone, (2) a novel C2f module as the model bottleneck that optimizes the traditional YOLO neck module for improving the model speed while maintaining similar performance, and (3) two segmentation heads that learn to predict the semantic segmentation masks for the input image. The YOLOv8 segmentation model has been shown to achieve state-of-the-art results on a wide variety of object detection and semantic segmentation benchmarks while maintaining high speed and efficiency.

\subsection{Evaluation Metrics}
This paper used a hybrid approach combining object-based and pixel-based metrics to assess segmentation performance. Object-based metrics assess segmentation success by measuring the intersections between predicted segment polygons and ground truth polygons. Pixel-based metrics assess the classification accuracy of individual pixels, including commonly used metrics such as IoU (Intersection over Union), Precision (P), and Recall (R). Throughout this article, true positives (TP), false positives (FP), and false negatives (FN) respectively refer to pixels correctly classified as belonging to the target class, pixels incorrectly classified as belonging to the target class, and pixels incorrectly classified as not belonging to the target class.

\subsubsection{IoU}

IoU, also known as the Jaccard Index, quantifies the overlap between the predicted pixels and the ground truth pixels for a target class. It ranges from 0 to 1, with higher values indicating better segmentation accuracy. 

\begin{equation}
    \begin{split}
    \textit{IoU} & = \frac{(\text{Prediction}) \cap (\text{Ground Truth})}{(\text{Prediction}) \cup (\text{Ground Truth})}  \\
    & = \frac{TP}{TP+FP+FN}
    \end{split}
\end{equation}

\subsubsection{Precision} 

Precision is the ratio of correctly predicted positive pixels to the total number of positive pixels. A high precision indicates that when the model predicts a positive pixel, it is likely to be correct. Precision is calculated by:

\begin{equation}
    \textit{Precision} = \frac{TP}{TP + FP}
\end{equation}

\subsubsection{Recall} 

Recall is the ratio of correctly predicted positive pixels to the total number of actual positive pixels. A high recall indicates that the model is effective at identifying all pixels of the positive class. Recall is calculated by:

\begin{equation}
    \textit{Recall} = \frac{TP}{TP + FN}
\end{equation}

\subsubsection{Object-based Metrics}

%\jz{any chance we can mention the word "topology" here with maybe one sentence? We claim topology metrics as our first contribution and I want to be brutally clear that this section is it. (the word topology currently has not been used again since the contribution)}\br{this brutal enough?}

Since the goal of segmentation is the identification of archaeological objects, a parallel series of object-based topological metrics were also developed to evaluate the inference results, which are perhaps more intuitively understood by non-DL specialists. The predicted masks of the inference data were converted to segmentation polygons for comparison with the ground truth polygons. The parity between the number of segmentation polygons and ground truth polygons provided a rough initial measure of success. Further analysis using GIS measured the intersections of segmentation polygons with ground truth polygons to evaluate segmentation success in a more detailed way. It was necessary to measure both the percentage of ground truth polygons intersecting inference polygons as well as the reverse since intersections were not always 1:1. That is, the areal overlap between a ground truth polygon and masks might be due to one or several mask polygons. Conversely, a given mask polygon might overlap several ground-truth polygons. %Results are provided in Section 4.3.

\section{Results}

\begin{figure*}[t]
  \centering
  \begin{subfigure}[t]{0.32\textwidth}
        \centering
        \includegraphics[width=\linewidth, height=0.7\linewidth]{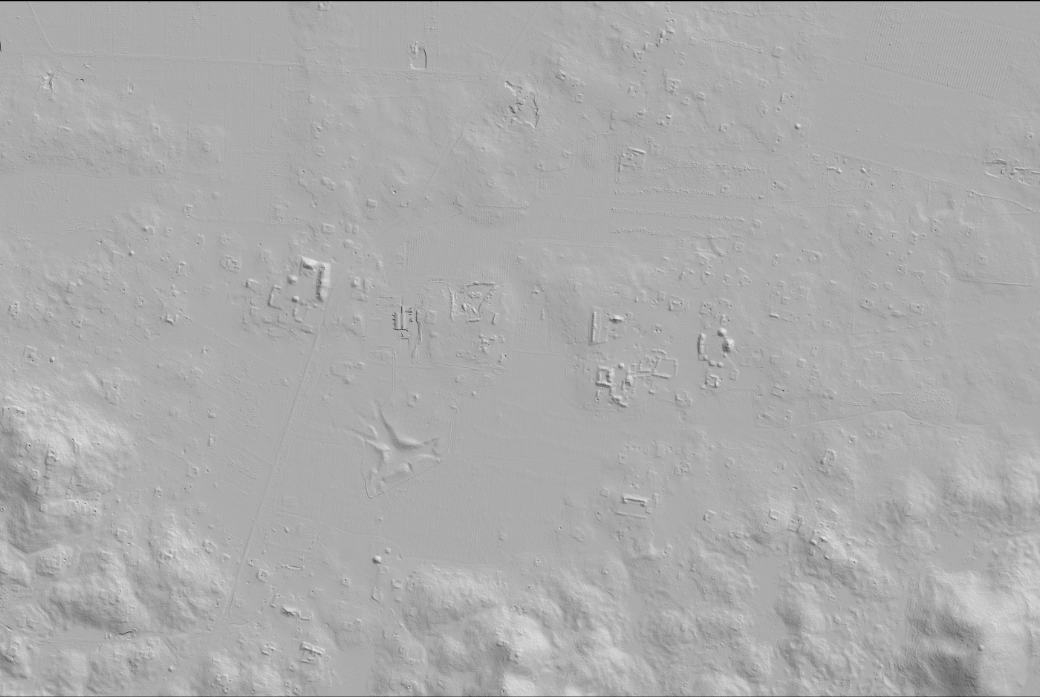}
        \caption{Muluchtzekel (MLS)}
        \end{subfigure}
   \hfill
   \begin{subfigure}[t]{0.32\textwidth}
        \centering
        \includegraphics[width=\linewidth, height=0.7\linewidth]{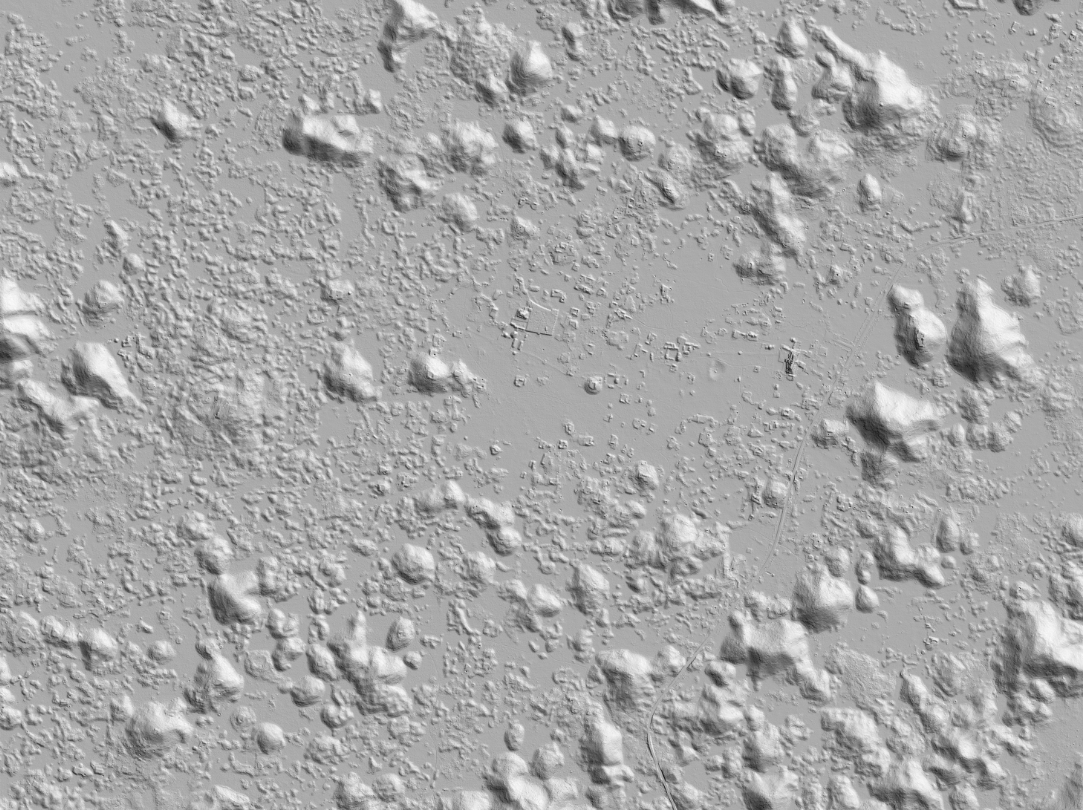}
        \caption{Sayil (SAY)}
        \end{subfigure}
   \hfill
   \begin{subfigure}[t]{0.32\textwidth}
        \centering
        \includegraphics[width=\linewidth, height=0.7\linewidth]{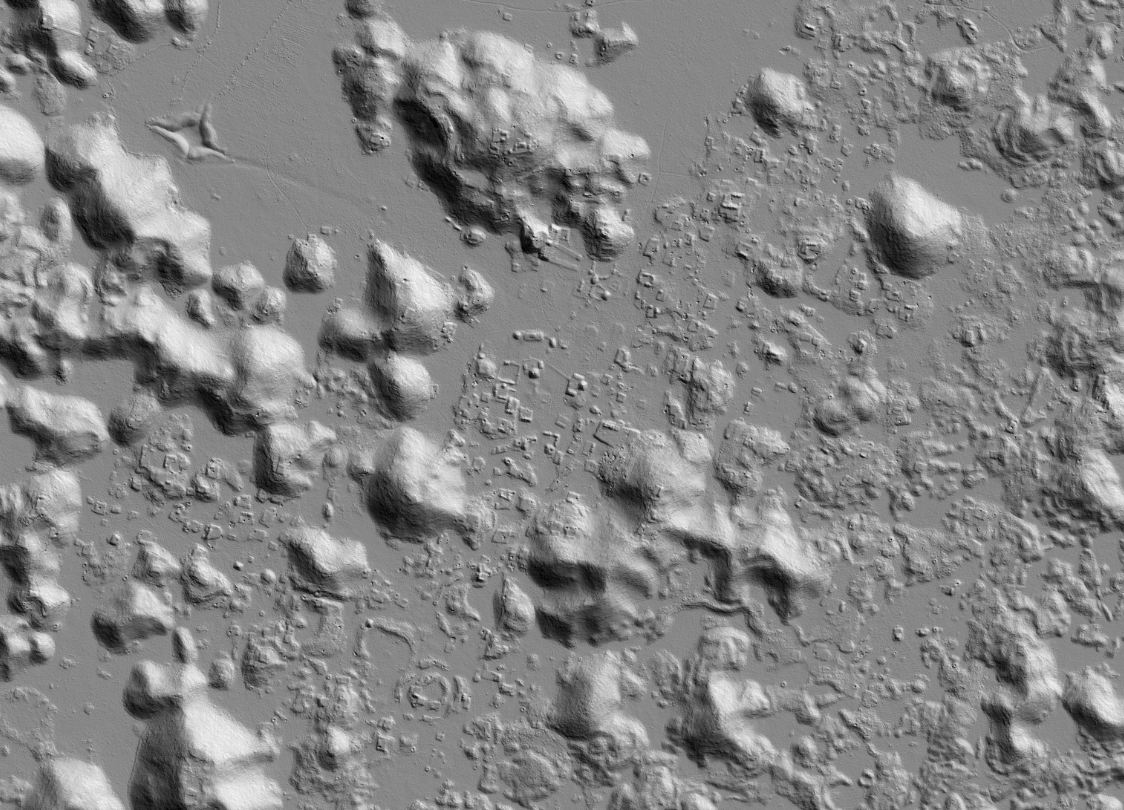}
        \caption{Huntichmul (HNT)}
        \end{subfigure}
   \caption{\label{fig:sites_HS}Hillshading image representation of MLS, SAY, and HNT sites where the "bumps" denote hill locations. The images of Muluchtzekel (MLS), Sayil (SAY), and Huntichmul (HNT) have a resolution of 6000$\times$4000, 6000 $\times$8000, and 3600$\times$5000 pixels respectively. The MLS and SAY sites are larger than the HNT despite being visualized in the same size. }
\end{figure*}

%\jz{Results are compiled across the three archaeological sites, each depicted with its corresponding hillshading representations as illustrated in Fig. \ref{fig:sites_HS}.} 

This section discusses the segmentation results across the three archaeological sites, Muluchtzekel (MLS), Sayil (SAY), and Huntichmul (HNT) (Table \ref{tab:tab-gnd-surv}, Fig. \ref{fig:sites_HS}).
%The LiDAR image of Muluchtzekel (MLS), Sayil (SAY), and Huntichmul (HNT) respectively has a resolution of 6000$\times$4000, 6000 $\times$8000, and 3600$\times$5000 pixels.} 
Data from MLS and SAY were utilized for training and testing the YOLOv8 model, while data from HNT were used as inference data to assess the model's effectiveness in a new site application. Performance improvements were realized as a combination of standard DL segmentation metrics, e.g., IoU, Recall, Precision, and a parallel GIS analysis. Ablation studies of this section demonstrate the performance impact of different pre-processing methods providing a best-practice recommendation.

%The YOLOv8 model was evaluated for the segmentation of two Maya structures, platforms, and annular structures. In the two archaeological sites for analysis (Muluchtzekel and Sayil), the house platform sample is significantly larger than that of annular structures, hence two image datasets were generated to avoid the data imbalance problem in deep learning. After the model was trained and tested on the customized datasets, the trained model was used to segment the two interested objects on a new site (Huntichmul), showcasing the potential of applying the proposed system to archaeological research. This was supplemented by a GIS analysis of the HNT results, evaluating the utility of the proposed approach in practice. An ablation study is then provided to demonstrate the effectiveness of the SPS visualization and scaling image in data augmentation.

Separate models were trained for platforms and annular structures. In both cases, computation was performed using an NVIDIA GeForce RTX 4090 GPU. The model to segment platforms was trained for 200 epochs with a batch size of 256. The model to segment annular structures was trained for 120 epochs also having batch size 256. No model parameters were fixed during training. Each dataset was divided into training (80\%), validation (10\%), and test (10\%) sets. During the training process, the model was trained on the training set and validated on the validation set. Following the completion of training, the trained model underwent evaluation using the test set.

\subsection{Testing on MLS and SAY}

%\begin{table*}
%    \centering
%    \begin{tabular}{cccccc}
%    \hline
%         & IoU & P & R & $\textbf{AP}_{50}$ & $\textbf{AP}_{50-95}$ \\
%    \hline
%    \hline
%        platform & 0.842 & 0.902 & 0.926 & 0.925 & 0.739 \\
        %platform \cite{bundzel2020semantic} & 0.720 & 0.725 & - & - & -\\
        %platform \cite{jannat2023extracting} & 0.74$^\ast$  & -  & - & - & - \\
%        \hline
%        annular & 0.809 & 0.891 & 0.898 & 0.91 & 0.479 \\
        %annular \cite{jannat2023extracting} & 0.82$^\ast$  & - & - & - & - \\
%    \hline
%    \end{tabular}
%    \caption{Statistical results of the platform and annular structure segmentation on the testing set.} % The trained model in this paper outperforms the state-of-the-art results in IoU and precision performance. $^\ast$Only performance on the validation set was reported in \cite{jannat2023extracting}.}
%    \label{tab:testing_results}
%\end{table*}

Table \ref{tab:test_and_infer_results} shows the evaluation metrics for the trained model on the testing data. The YOLOv8 model trained for platform segmentation achieved an IoU performance of 0.842, with precision and recall rates exceeding 0.9. The annular structure model attained an IoU performance of 0.809, with precision and recall rates nearing 0.9. These results reflect high accuracy in locating and segmenting platforms and annular structures. The difference in performance between these two object types can be attributed to (1) the much smaller size of annular structures when compared to platforms, as shown in Fig. \ref{fig:hist_area}, making falsely segmented pixels have a relatively greater impact on the evaluation metrics for annular structures, and (2) the substantially fewer data instances in the annular structure dataset compared to the platform dataset, posing a challenge for the YOLOv8 model to learn features as effectively as it did for platforms. Thus, the overall performance on platforms benefits from distinct image features and a larger volume of training data than annular structures. %, which contributed to enhanced generalization and robustness. 

\begin{table}
    \centering
    \begin{tabular}{ccccc}
    \hline
        & & IoU & P & R  \\
    \hline
    \hline
        \multirow{2}{6em}{MLS \& SAY (testing)} & platform & 0.842 & 0.902 & 0.926  \\
        %platform \cite{bundzel2020semantic} & 0.720 & 0.725 & - \\
        %platform \cite{jannat2023extracting} & 0.74$^\ast$  & -  & - \\
        
        & annular & 0.809 & 0.891 & 0.898 \\
        %annular \cite{jannat2023extracting} & 0.82$^\ast$  & - & - \\
    \hline
        \multirow{2}{6em}{HNT (inference)} & platform & 0.604 & 0.718 & 0.792 \\
        & annular & 0.537 & 0.611 & 0.817 \\
    \hline
    \end{tabular}
    \caption{Statistical results of the platform and annular structure segmentation on the testing set.} % The trained model in this paper outperforms the state-of-the-art results in IoU and precision performance. $^\ast$Only performance on the validation set was reported in \cite{jannat2023extracting}.}
    \label{tab:test_and_infer_results}
\end{table}

Previous research in ancient Maya structure segmentation had also reported results regarding platforms and, in one case, annular structures. The state-of-the-art performance includes achieving a 0.72 IoU score for platform segmentation on a dataset containing approximately 1900 platform samples from Petén, Guatemala \cite{bundzel2020semantic}, and a 0.765 IoU score on a dataset consisting of 952 image tiles, each containing at least one platform from Chactún, Mexico \cite{ayala2022deep}. Additionally, authors in \cite{jannat2023extracting} reported an IoU score of 0.74 for platform segmentation and 0.82 for annular structures in their validation set. Due to the unavailability of these Maya datasets, direct comparison with our work is not possible at this time. However, the results of our study indicate that the proposed pre-process, data augmentation and post-processing techniques can contribute to improving performance for both platforms and annular structures.

\subsection{Inference on HNT}

Table \ref{tab:test_and_infer_results} shows the evaluation metrics for the trained model on a third site, Huntichmul (HNT), that contains 513 platform targets and 70 annular structures. The overall IoU score was 0.604 for platform segmentation and 0.537 for the annular structure, both having an approximately 0.25 decrease from the performance on the MLS and SAY sites. The precision and recall results shared a similar decrease as IoU. One possible explanation for this difference in performance could be the relatively smoother terrain (more flats and low rises) present in the training data (MLS and SAY) compared to the inference data (HNT), as indicated by the distribution of terrain types in Table \ref{tab:tab-gnd-surv}. The performance of HNT may be improved by incorporating additional training samples from rugged terrain.

%\begin{table}
%    \centering
%    \begin{tabular}{ccccccc}
%    \hline
%         & & GT & pred & IoU & P & R \\
%    \hline
%    \hline
%        \multirow{5}{3em}{platform} & all & 513 & 493 & 0.604 & 0.718 & 0.792 \\
%        & area A & 75 & 69 & 0.706 & 0.814 & 0.842  \\
%        & area B & 142 & 126 & 0.655 & 0.787 & 0.797 \\
%        & area C & 43 & 38 & 0.719 & 0.772 & 0.912 \\
%        & area D & 32 & 36 & 0.685 & 0.758 & 0.877 \\
%    \hline
%        \multirow{5}{3em}{annular} & all & 70 & 82 & 0.537 & 0.611 & 0.817 \\
%        & area E & 22 & 22 & 0.636 & 0.709 & 0.860  \\
%        & area F & 13 & 13 & 0.649 & 0.745 & 0.834 \\
%        & area G & 10 & 12 & 0.642 & 0.733 & 0.837 \\
%        & area H & 11 & 14 & 0.472 & 0.559 & 0.753 \\
%    \hline
%    \end{tabular}
%    \caption{Statistical inference results on HNT. \jz{analysis on the topology (elevation per unit in each region) and add regions that not performing very well too. The 4 regions for platform are \{2000:3000, 1200:2200\}, \{1800:3000, 700:2500\}, \{1623:2144, 530:1567\}, \{423:1593, 1171:1675\}. The 4 regions for annular are \{2984:5001, 1777:3601\}, \{537:1766, 1897:3378\}, \{1:957, 1:1831\}, \{1029:1904, 128:2113\}}}
%    \label{tab:infer_results_on_HNT}
%\end{table}

\begin{figure}
    \centering
    \includegraphics[width=0.85\linewidth]{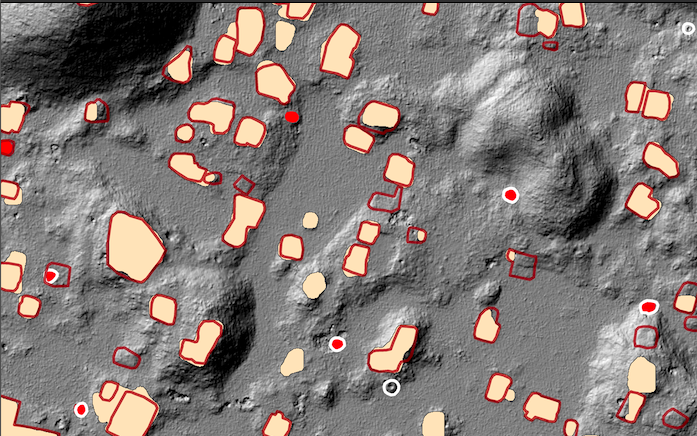}
    \caption{An excerpt of the inference result of HNT where the brown and white polygons respectively indicate the ground truth boundaries of platforms and annular structures and the solid beige and red regions respectively show the predictions of YOLOv8.}
    \label{fig:HNT_seg_result_close_up}
\end{figure}

Figures \ref{fig:HNT_seg_result} and \ref{fig:HNT_seg_result_close_up} illustrate excerpts of the inference results obtained from HNT. Fig. \ref{fig:HNT_seg_result} offers a broader field of view, while Fig. \ref{fig:HNT_seg_result_close_up} provides a closer view, allowing for a more detailed examination of the results. The segmentation results showcase success in accurately identifying the majority of the objects, with failures primarily observed on smaller targets.

\subsection{Topology Analyses of HNT Results}

Table \ref{tab:polygon_results} presents the results of the object-based topology analyses. Approximately 83\% of platform and annular ground truth polygons intersected an inference polygon, while 79.5\% of platform and 70.7\% of annular segmentation polygons intersected ground truth polygons, the difference perhaps attributable to the underproduction of inference platforms and overproduction of inference annular structures.

%Table \ref{tab:polygon_results} shows the percentage of intersections between the ground truth (GT) polygons and the predicted polygons generated by YOLOv8 at the HNT site. These intersections are measured both within the ground truth polygons and within the predicted polygons. Using GIS, the YOLOv8 raster masks were converted to polygons which were then compared to the labeled polygons. GIS enables the measurement of the number of intersections between labeled and YOLOv8 polygons, as well as the actual area of overlap. It is important to measure both the percentage of labeled polygons intersecting YOLOv8 polygons as well as the reverse to obtain more understanding of how well a segmentation performed in practice. As demonstrated in Table \ref{tab:polygon_results}, YOLOv8 prediction intersects with over 80\% of the polygons in the ground truth (GT) for both platforms and annular structures. Specifically, approximately 80\% of platform polygons in the prediction intersect with the ground truth, while the figure stands at 70\% for annular structures. These results showcase YOLOv8's effectiveness in identifying Maya structures. Smaller objects like annular structures pose a greater challenge for YOLOv8 to accurately segment, resulting in a lower intersection rate compared to platforms. Intersection over ground truth and intersection over prediction, as two of a series of object/polygon-based metrics paralleling the pixel-based metrics discussed in previous sections, are perhaps more intuitively appreciated by the non-specialist. 

\begin{table}[h]
    \centering
    \begin{tabular}{cccccc}
    \hline
        & & Total polygons & Intersects & Pct.\\
    \hline
    \hline
		\multirow{2}{4em}{platform} & GT & 513 & 426 & 0.830 \\
		 & pred & 478 & 380 & 0.795 \\
   \hline
        \multirow{2}{4em}{annular} & GT &  70 & 58 &  0.829 \\
        & pred & 82 & 58 & 0.707 \\
    \hline
    \end{tabular}
    \caption{The percentage (Pct.) of intersections of the ground truth (GT) and YOLOv8 predicted polygons from HNT respectively in the ground truth (GT) and YOLOv8 predicted polygons.}  
    \label{tab:polygon_results}
\end{table}

Table \ref{tab:quartile_results} divides intersection successes into quartiles based on the area of the ground truth polygons. This shows a clear contrast between the high number of failures in the lowest quartile, with less than 60\% successful, with the very robust statistics of the upper three quartiles, with average success rates of 91.4\% for platforms and 94.1\% for annular structures. Reasons for the failure of small polygons to segment correctly include (1) their resemblance to natural features, (2) the "bumpiness" of the LiDAR data, and (3) the small number of pixels involved.

\begin{table}
    \centering
    \begin{tabular}{cccc}
    \hline
        & Quartile & Max. Area ($m^2$) & Intersect Pct.\\
    \hline
    \hline
        \multirow{4}{4em}{platform} & 1 & 210.7 & 0.581\\
         & 2 & 359.5 & 0.852\\
         & 3 & 633.3 & 0.906\\
         & 4 & 5700.3 & 0.984\\
         & overall & 5700.3 & 0.830\\  
         \hline
        \multirow{4}{4em}{annular} & 1 & 74.0 & 0.500\\
         & 2 & 103.6 & 0.882\\
         & 3 & 125.5 & 0.941\\
         & 4 & 238.2 & 1.0\\
         & overall & 238.2 & 0.829\\ 
    \hline
    \end{tabular}
    %\caption{Percentages of overlaps between HNT ground truth and YOLOv8-predicted masks in different quartiles of area. Objects of interest were split into 4 quartiles based on their area with Quartile 1 being the smallest group and Quartile 4 the largest.}
    \caption{Percentages of the HNT segmented polygons that intersect the ground truth polygons. The ground truth polygons were sorted by area and divided into quartiles to investigate where failures were occurring.}  

    \label{tab:quartile_results}
\end{table}

\subsection{Ablation Studies}

This section provides results for two ablation studies that investigate the effectiveness of (1) the SPS representation of the data compared to other representations and (2) the impact of multi-scale methods for training and inference on our data. Both studies evaluate performance using the platform dataset which has more samples and therefore can be expected to yield more accurate results for these studies. We feel that the conclusions drawn from the ablation studies generalize across segmentation tasks for other objects within the dataset. 

\subsubsection{Airborne LiDAR Data Representation}

\begin{figure}
    \centering
    \begin{subfigure}[t]{0.145\textwidth}
        \centering
        \includegraphics[width=\linewidth]{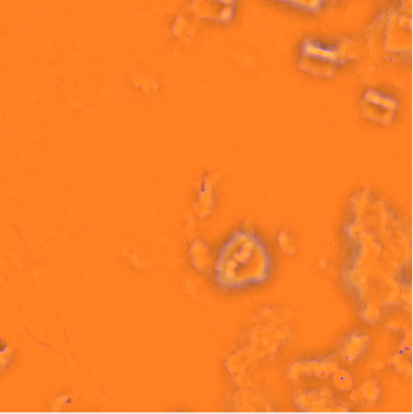}
        \caption{SPS}
    \end{subfigure}
    \hfill
    \begin{subfigure}[t]{0.145\textwidth}
        \centering
        \includegraphics[width=\linewidth]{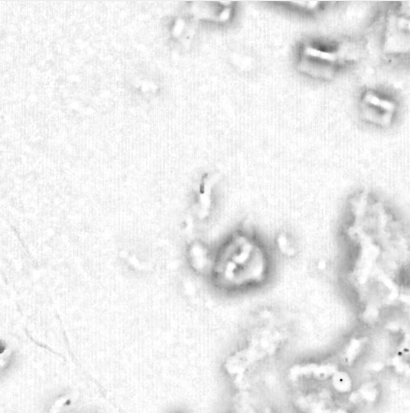}
        \caption{SVF}
    \end{subfigure}
    \hfill
    \begin{subfigure}[t]{0.145\textwidth}
        \centering
        \includegraphics[width=\linewidth]{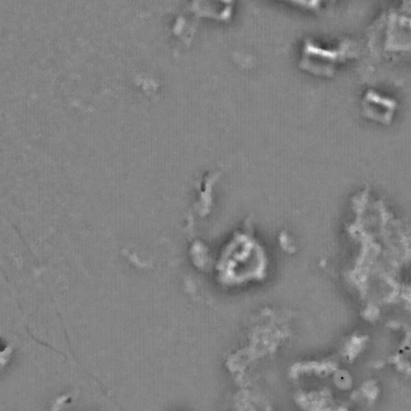}
        \caption{PO}
    \end{subfigure}
    \\
    \begin{subfigure}[t]{0.145\textwidth}
        \centering
        \includegraphics[width=\linewidth]{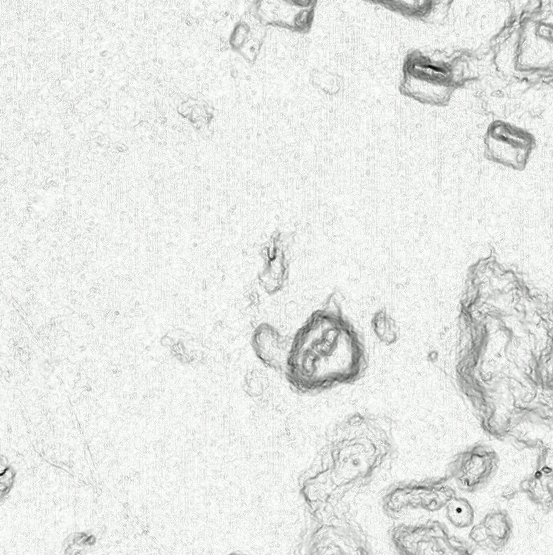}
        \caption{slope}
    \end{subfigure}
    \hfill
    \begin{subfigure}[t]{0.145\textwidth}
        \centering
        \includegraphics[width=\linewidth]{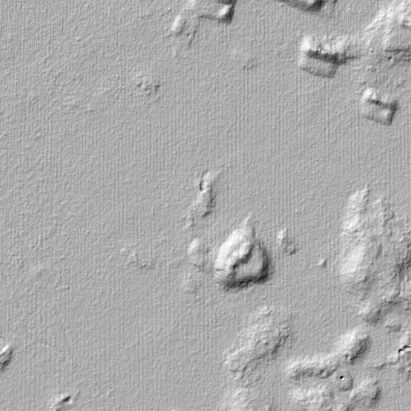}
        \caption{HS}
    \end{subfigure}
    \hfill
    \begin{subfigure}[t]{0.145\textwidth}
        \centering
        \includegraphics[width=\linewidth]{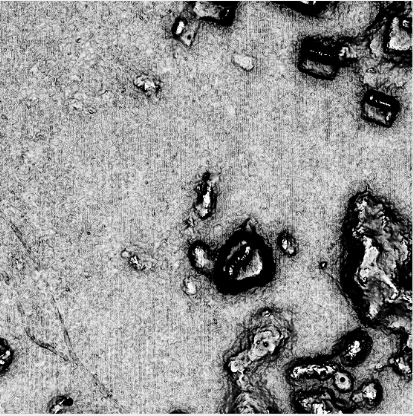}
        \caption{elevation}
    \end{subfigure}
\caption{Different ALS data representations.\label{fig:als_vis}}
\end{figure}

Table \ref{tab:results_for_diff_vis} presents the performance of platform segmentation using different ALS data representations derived from the DTM. The elevation data within each training tile was normalized to the $[0,1]$ interval, a practice known to enhance classification outcomes, as corroborated by prior studies \cite{hliboky2022detecting, jannat2023extracting}. Examples of these representations are depicted in Fig. \ref{fig:als_vis}. To create three-channel data for training input, the single-channel slope, PO, SVF, HS, and elevation data were replicated across the remaining two channels. 

\begin{table}[h]
    \centering
    \begin{tabular}{cccccc}
    \hline
         & IoU & P & R \\
    \hline
    \hline
        SPS & \textbf{0.842} & 0.902 & \textbf{0.926} \\
        SVF & 0.832 & 0.899 & 0.917 \\
        PO & 0.838 & \textbf{0.903} & 0.923 \\
        slope & 0.823 & 0.901 & 0.905 \\
        HS & 0.834 & 0.901 & 0.918 \\
        elevation & 0.812 & 0.889 & 0.904 \\
    \hline
    \end{tabular}
    \caption{Statistical testing results of platform segmentation from using different ALS data representations.}
    \label{tab:results_for_diff_vis}
\end{table}

As indicated in Table \ref{tab:results_for_diff_vis}, employing SPS yields better results in terms of IoU score and recall rate, ranking second in precision rate with a marginal difference of 0.001 from the top performer. SPS, a composite of the slope, PO, and SVF data, offers richer information compared to individual data components, potentially furnishing more discernible features for DL networks to exploit. On the other hand, employing elevation data and slope data (the first gradient of elevation) trails behind in all three metrics, suggesting that elevation data alone may not suffice for a deep learning system to effectively learn the segmentation of Maya structures. While hillshading and elevation representations may seem visually appealing to the eye and could provide researchers with an intuitive understanding when examining archaeological sites, this does not necessarily translate to improved performance for DL segmentation models.

\subsubsection{Multi-scale}

Table \ref{tab:results_for_diff_scale} shows the performance of testing and inference when incorporating multi-scale methods images into data augmentation training and inference post-processing. Two networks were trained: one with scaling the training data (``\textit{HNT w/o scaling''}) and the other without scaling (``\textit{train w/o scaling''}). These networks were evaluated in three scenarios: (1) on the test set with the same scaling factor as the training set (\textit{``test w/ scaling''} and \textit{``test w/o scaling''}), (2) on inferring the HNT image (\textit{``HNT w/o scaling''}), and (3) on inferring a combination of the HNT image and its scaled variant (\textit{``HNT w/ scaling''}) using the approach described in Section \ref{sec:infer}. 

As demonstrated in Table \ref{tab:results_for_diff_scale}, the best testing outcomes are observed when the training data incorporate scaled data. Similarly, the best inference results occur when the network is trained with scaled data and applied to infer the HNT combination. Overall, the results in \textit{``HNT w/o scaling''} are better than those in \textit{``train w/o scaling''}. Notably, scaling the HNT image leads to a notable improvement in inference performance, enhancing results by approximately 8\% for both networks trained with and without scaled data.

\section{End-user Benefits}

The benefits of image segmentation for the archaeologist are clear. Manual identification of platforms in a large LiDAR dataset can cost weeks or months of dedicated analyst time. Consistency in identification may also be an issue, and with larger datasets, there is rarely the luxury of full re-analysis. Platform polygons vary tremendously in their footprints, while small natural undulations in the terrain can easily be mistaken for construction. Thus, image segmentation provides the end-user with a rapid and consistent initial pass through the LiDAR dataset, providing reliable identifications for a high proportion of the larger structures. Human analysts can then devote added time to smaller and more ambiguous features. %In contrast, over 80\% of the test platforms and annular structures from MLS and SAY were identified by YOLOv8. In the application of the trained network to a new site HNT, if the lowest quartile by area is excluded, a remarkable 91.4\% of actual test polygons were identified by YOLOv8. Similar results were obtained for annular structures, except that the identification rate of the top three quartiles climbed to 94.1\%.

%Nevertheless, image segmentation is not yet at the point where it is superior to human judgment. 

\begin{table}[h]
    \centering
    \begin{tabular}{ccccc}
    \hline
        & & IoU & P & R \\
    \hline
    \hline
        \multirow{3}{4em}{train w/ scaling} & test w/ scaling & 0.842 & 0.902 & 0.926 \\
        & HNT w/o scaling & 0.559 & 0.804 & 0.647 \\
        & HNT w/ scaling & 0.604 & 0.718 & 0.792 \\
    \hline
        \multirow{3}{4em}{train w/o scaling} & test w/o scaling & 0.83 & 0.894 & 0.921 \\
        & HNT w/o scaling & 0.549 & 0.807 & 0.632 \\
        & HNT w/ scaling & 0.591 & 0.741 & 0.745 \\
    \hline
    \end{tabular}
    \caption{Statistical testing and HNT inference results of platform segmentation when applying different scale factors to generating the training data.}
    \label{tab:results_for_diff_scale}
\end{table}

As to the ability of segmentation to accurately determine platform outlines, the results are often remarkably close to what a human observer would posit. Given the variety of footprints, however, it is probably premature to invest too much faith in their accuracy. Human analysts will be able to accept many segmentation masks but will also need to manually adjust a substantial percentage.

\section{Conclusion}

This study automates the identification and segmentation of archaeological structures, such as annular structures and platforms, showcasing the immense potential of deep learning in streamlining labor-intensive tasks traditionally reliant on manual expertise. By introducing a novel data processing pipeline and achieving state-of-the-art segmentation performances, this research not only pushes the boundaries of deep learning applications in archaeology but also addresses the unique challenges posed by aerial LiDAR data. The utilization of YOLOv8 not only enhances the accuracy of archaeological structure segmentation but also presents a promising avenue for the automated analysis of vast archaeological landscapes. As these deep learning technologies continue to advance, their potential to revolutionize Maya object recognition in large LiDAR datasets is increasingly evident, promising new insights and discoveries that contribute to the broader field of archaeology.

\section*{Acknowledgments}

We would like to acknowledge funding for LiDAR acquisition from the National Science Foundation (Award No. 1660503) and the National Center for Airborne LiDAR Mapping for the collection of the data used in this article.

{
\small
\bibliographystyle{ieeenat_fullname}
\bibliography{main}
}


% \clearpage
% \setcounter{page}{1}
% \maketitlesupplementary

% \section{Rationale}
% \label{sec:rationale}
% % 
% Having the supplementary compiled together with the main paper means that:
% % 
% \begin{itemize}
% \item The supplementary can back-reference sections of the main paper, for example, we can refer to \cref{sec:intro};
% \item The main paper can forward reference sub-sections within the supplementary explicitly (e.g. referring to a particular experiment); 
% \item When submitted to arXiv, the supplementary will already included at the end of the paper.
% \end{itemize}
% % 
% To split the supplementary pages from the main paper, you can use \href{https://support.apple.com/en-ca/guide/preview/prvw11793/mac#:~:text=Delete%20a%20page%20from%20a,or%20choose%20Edit%20%3E%20Delete).}{Preview (on macOS)}, \href{https://www.adobe.com/acrobat/how-to/delete-pages-from-pdf.html#:~:text=Choose%20%E2%80%9CTools%E2%80%9D%20%3E%20%E2%80%9COrganize,or%20pages%20from%20the%20file.}{Adobe Acrobat} (on all OSs), as well as \href{https://superuser.com/questions/517986/is-it-possible-to-delete-some-pages-of-a-pdf-document}{command line tools}.

\end{document}